\documentclass[10pt,twocolumn,letterpaper]{article}
\usepackage[accsupp]{axessibility}
\usepackage{cvpr}              

\usepackage{graphicx}
\usepackage{amsmath}
\usepackage{amssymb}
\usepackage{booktabs}
\usepackage{multirow}
\usepackage{makecell}
\usepackage{bbm}
\usepackage{comment}

\usepackage[pagebackref,breaklinks,colorlinks]{hyperref}

\usepackage[capitalize]{cleveref}
\crefname{section}{Sec.}{Secs.}
\Crefname{section}{Section}{Sections}
\Crefname{table}{Table}{Tables}
\crefname{table}{Tab.}{Tabs.}

\def\cvprPaperID{5778} 
\def\confName{CVPR}
\def\confYear{2023}

\newcommand{\yankun}[1]{\textcolor{red}{#1}}

\begin{document}

\title{Two-shot Video Object Segmentation}

\author{Kun Yan$^{1}$\quad Xiao Li$^{2}$  \quad Fangyun Wei$^{2}$\quad Jinglu Wang$^{2}$ \quad Chenbin Zhang$^{1}$ \quad Ping Wang$^1$\footnotemark[1] \quad Yan Lu$^{2}$\footnotemark[1]\thanks{Corresponding authors.} \vspace{4pt}\\
	$^1$Peking University \quad
    $^2$Microsoft Research Asia \\
    {\tt\small \{kyan2018, zcbin, pwang\}@pku.edu.cn} \quad
	{\tt\small \{xili11, fawe, jinglwa, yanlu\}@microsoft.com} \\
}

\maketitle

\begin{abstract}
Previous works on video object segmentation~(VOS) are trained on densely annotated videos. Nevertheless, acquiring annotations in pixel level is expensive and time-consuming. In this work, we demonstrate the feasibility of training a satisfactory VOS model on sparsely annotated videos—we merely require two labeled frames per training video while the performance is sustained. We term this novel training paradigm as two-shot video object segmentation, or two-shot VOS for short. The underlying idea is to generate pseudo labels for unlabeled frames during training and to optimize the model on the combination of labeled and pseudo-labeled data. Our approach is extremely simple and can be applied to a majority of existing frameworks. We first pre-train a VOS model on sparsely annotated videos in a semi-supervised manner, with the first frame always being a labeled one. Then, we adopt the pre-trained VOS model to generate pseudo labels for all unlabeled frames, which are subsequently stored in a pseudo-label bank. Finally, we retrain a VOS model on both labeled and pseudo-labeled data without any restrictions on the first frame. For the first time, we present a general way to train VOS models on two-shot VOS datasets. By using $7.3\%$ and $2.9\%$ labeled data of YouTube-VOS and DAVIS benchmarks, our approach achieves comparable results in contrast to the counterparts trained on fully labeled set. Code and models are available at \url{https://github.com/yk-pku/Two-shot-Video-Object-Segmentation}. 
\end{abstract}

\section{Introduction}
\label{sec:intro}

\begin{figure}
  \centering
  \begin{subfigure}{1.0\linewidth}
  \centering
    \includegraphics[width=1.0\columnwidth]{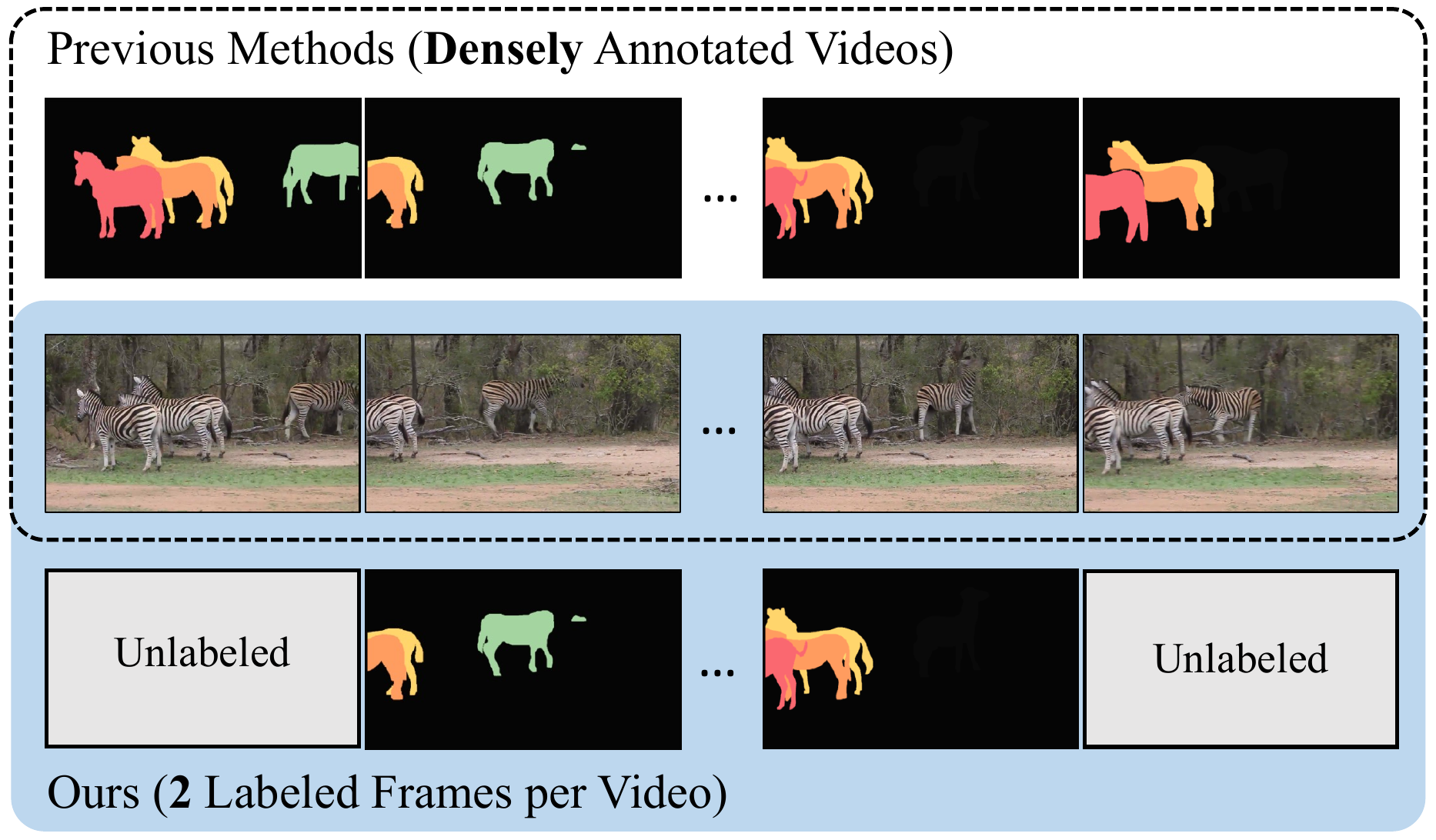}
    \caption{Previous works on video object segmentation rely on densely annotated videos. We present two-shot video object segmentation, which merely accesses two labeled frames per video.}
    \label{fig:teaser-a}
  \end{subfigure}
  \hfill
  \begin{subfigure}{1.0\linewidth}
  \centering
    \includegraphics[width=1.0\columnwidth]{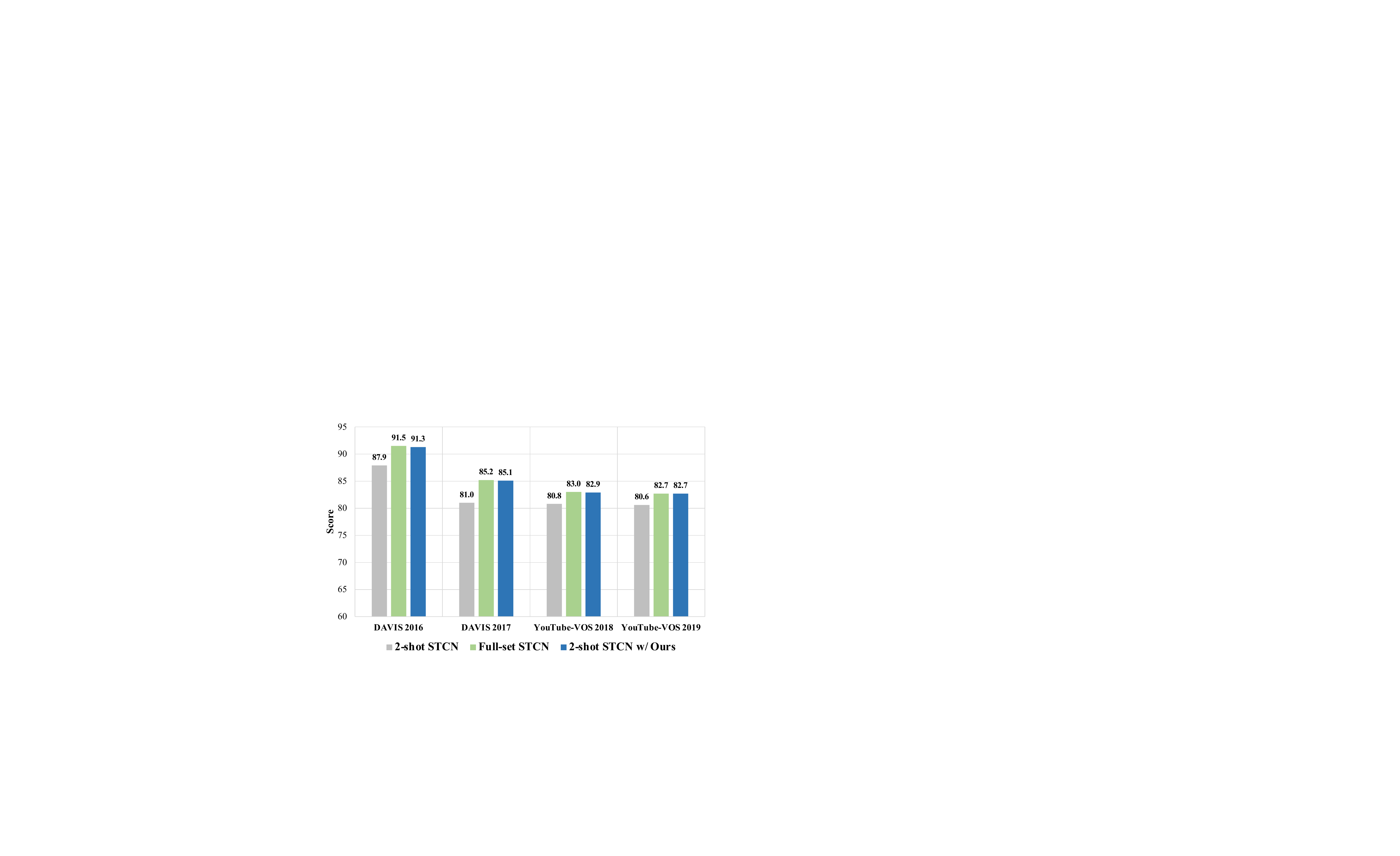}
    \caption{Comparison among naive 2-shot STCN, STCN trained on full set and 2-shot STCN equipped with our approach on DAVIS 2016/2017 and YouTube-VOS 2018/2019. 
    }
    \label{fig:teaser-b}
  \end{subfigure}
  \caption{(a) Problem formulation. (b) Comparison among STCN variants on various datasets.}
  \label{fig:teaser}
\end{figure}

Video object segmentation (VOS), also known as mask tracking, aims to segment the target object in a video given the annotation of the reference (or first) frame. Existing approaches~\cite{oh2019video,yang2020collaborative,seong2021hierarchical,cheng2021rethinking,xie2021efficient,li2022recurrent,cheng2022xmem} are trained on densely annotated datasets such as DAVIS~\cite{perazzi2016benchmark,pont20172017} and YouTube-VOS~\cite{xu2018youtube}. However, acquiring dense annotations, particularly at the pixel level, is laborious and time-consuming. For instance, the DAVIS benchmark consists of 60 videos, each with an average of 70 labeled frames; the YouTube-VOS dataset has an even larger amount of videos, and every fifth frame of each video is labeled to lower the annotation cost. It is necessary to develop data-efficient VOS models to reduce the dependency on labeled data.

In this work, we investigate the feasibility of training a satisfactory VOS model on sparsely annotated videos. For the sake of convenience, we use the term $N$-shot to denote that $N$ frames are annotated per training video. Note that $1$-shot is meaningless since it degrades VOS to the task of image-level segmentation. We use STCN~\cite{cheng2021rethinking} as our baseline due to its simplicity and popularity. Since at least two labeled frames per video are required for VOS training, we follow the common practice to optimize a naive $2$-shot STCN model on the combination of YouTube-VOS and DAVIS, and evaluate on YouTube-VOS 2018/2019 and DAVIS 2016/2017, respectively. We compare the native $2$-shot STCN with its counterpart trained on full set in~\cref{fig:teaser-b}. Surprisingly, 2-shot STCN still achieves decent results, for instance, only a $-2.1\%$ performance drop is observed on YouTube-VOS 2019 benchmark, demonstrating the practicality of $2$-shot VOS.

So far, the wealth of information present in unlabeled frames is yet underexplored. In the last decades, semi-supervised learning, which combines a small amount of labeled data with a large collection of unlabeled data during training, has achieved considerable success on various tasks such as image classification~\cite{sohn2020fixmatch,berthelot2019mixmatch}, object detection~\cite{sohn2020simple,xu2021end} and semantic segmentation~\cite{hu2021semi,ke2020guided}. In this work, we also adopt this learning paradigm to promote $2$-shot VOS (see~\cref{fig:teaser-a}). The underlying idea is to generate credible pseudo labels for unlabeled frames during training and to optimize the model on the combination of labeled and pseudo-labeled data. 
Here we continue to use STCN~\cite{cheng2021rethinking} as an example to illustrate our design principle, nevertheless, our approach is compatible with most VOS models. Concretely, STCN takes a randomly selected triplet of labeled frames as input but the supervisions are only applied to the last two—VOS requires the annotation of the first frame as reference to segment the object of interest that appeared in subsequent frames. This motivates us to utilize the ground-truth for the first frame to avoid error propagation during early training. Each of the last two frames, nevertheless, can be either a labeled frame or an unlabeled frame with a high-quality pseudo label. 
Although the performance is improved with this straightforward paradigm, the capability of semi-supervised learning is still underexplored due to the restriction of employing the ground truth as the starting frame. 
We term the process described above as \textit{phase-1}.

To take full advantage of unlabeled data, we lift the restriction placed on the starting frame, allowing it to be either a labeled or pseudo-labeled frame. To be specific, we adopt the VOS model trained in \textit{phase-1} to infer the unlabeled frames for pseudo-labeling. After that, each frame is associated with a pseudo label that approximates the ground-truth. The generated pseudo labels are stored in a pseudo-label bank for the convenience of access. The VOS model is then retrained without any restrictions—similar to
how it is trained through supervised learning, but each frame has either a ground-truth or a pseudo-label attached to it. It is worth noting that, as training progresses, the predictions become more precise, yielding more reliable pseudo labels—we update the pseudo-label bank once we identify such pseudo labels. 
The above described process is named as \textit{phase-2}. 
As shown in \cref{fig:teaser-b}, our approach assembled onto STCN, achieves comparable results (\textit{e.g.} 85.2\% v.s 85.1\%  on DAVIS 2017, and 82.7\%  v.s 82.7\%  on YouTube-VOS 2019) in contrast to its counterpart, STCN trained on full set, though our approach merely accesses 7.3\% and 2.9\% labeled data of YouTube-VOS and DAVIS benchmark, respectively.

Our contributions can be summarized as follows:
\begin{itemize}
    \item For the first time, we demonstrate the feasibility of two-shot video object segmentation: two labeled frames per video are almost sufficient for training a decent VOS model, even without the use of unlabeled data.
    \item We present a simple yet efficient training paradigm to exploit the wealth of information present in unlabeled frames. This novel paradigm can be seamlessly applied to various VOS models, \textit{e.g.}, STCN~\cite{cheng2021rethinking}, RDE-VOS~\cite{li2022recurrent} and XMem~\cite{cheng2022xmem} in our experiments.
    \item  Though we only access a small amount of labeled data (\textit{e.g.} $7.3\%$ for YouTube-VOS and $2.9\%$ for DAVIS), our approach still achieves competitive results in contrast to the counterparts trained on full set. For example, 2-shot STCN equipped with our approach achieves 85.1$\%$/82.7$\%$ on DAVIS 2017/YouTube-VOS 2019, which is +4.1$\%$/+2.1$\%$ higher than the naive 2-shot STCN while -0.1$\%$/-0.0$\%$ lower than the STCN trained on full set.
\end{itemize}

\section{Related work}
\noindent\textbf{Video object segmentation.}
Existing VOS methods can be categorized into two groups: online-learning methods and offline-learning methods. Online-learning methods~\cite{caelles2017one,luiten2018premvos,perazzi2017learning,voigtlaender2017online,xiao2018monet,maninis2018video,cheng2017segflow} need to fine-tune the networks at test time based on the query mask of the first frame. However, test-time fine-tuning is computationally expensive. 
In contrast, offline-learning methods~\cite{mao2021joint,yang2021associating,hu2021learning,zhang2020transductive,ge2021video,lu2020video} aim at training a model that segments videos without any adaptations during inference. It is usually achieved via propagation and matching. Propagation-based methods~\cite{chen2020state,li2018video,oh2018fast,johnander2019generative} segment the target object sequentially by propagating the reference mask of the first frame. Matching-based methods~\cite{cheng2021rethinking,yang2020collaborative,oh2019video,wang2021swiftnet} typically employ a memory bank to store the features of a collection of frames, then a feature matching is adopted to segment the query frame.

STM~\cite{oh2019video} received widespread attention among the matching-based methods. STM proposes to construct a memory network to store the masks of the previous frames. Then the query frame is segmented using the information stored in the memory. A majority of follow-up works improved STM in several aspects~\cite{seong2020kernelized,seong2021hierarchical,cheng2022xmem,li2022recurrent,xie2021efficient}. 
For example, STCN~\cite{cheng2021rethinking} establishes correspondences between frames to avoid re-encoding the mask feature of each object; RDE-VOS~\cite{li2022recurrent} builds a constant-size memory bank by recurrent dynamic embedding while retaining the performance; XMem~\cite{cheng2022xmem} incorporates multiple feature memory stores and achieves the best performance. 
Despite their promising results, these methods need densely annotated videos for training. Instead, our method only needs two labeled frames per video and is compatible with most VOS models. It is worth noting that the meaning of ``one-shot'' claimed by \cite{caelles2017one} significantly differs from that of our ``two-shot''. In \cite{caelles2017one}, ``one-shot'' refers to that given a reference frame during inference, the optimized model is able to segment the remaining frames. In contrast, we use the term ``$N$-shot'' to denote the number of labeled frames per video. Therefore, in our setting, ``one-shot'' denotes that only a single labeled frame per video is available during training.

\noindent\textbf{Semi-supervised learning.}
Semi-supervised learning is an efficient way to improve model performance by using a few labeled data and a large amount of unlabeled data. It has achieved promising results across various computer vision tasks, such as image classification~\cite{sohn2020fixmatch,tarvainen2017mean}, image segmentation~\cite{hu2021semi,ke2020guided}, object detection~\cite{sohn2020simple,xu2021end} and action recognition~\cite{xu2022cross}. 
The dominated works can be roughly categorized into consistency based methods~\cite{laine2016temporal,tarvainen2017mean,berthelot2019mixmatch,sajjadi2016regularization,french2019semi,chen2021semi} and pseudo-labeling based methods~\cite{lee2013pseudo,sohn2020simple,zoph2020rethinking,xie2020self,grandvalet2004semi,zou2020pseudoseg}. Consistency based methods enforce consistency between predictions of different perturbations, such as model perturbing~\cite{bachman2014learning}, data augmentations~\cite{xie2020unsupervised,berthelot2019remixmatch} and adversarial perturbations~\cite{miyato2018virtual}. Pseudo-labeling based methods generate one-hot pseudo labels for unlabeled data. Then the model is optimized on the combination of labeled data and pseudo-labeled data. Our approach also adopts pseudo-labeling to improve two-shot VOS.

\section{Methodology}
\begin{figure*}[t]
  \centering
   \includegraphics[width=0.85\textwidth]{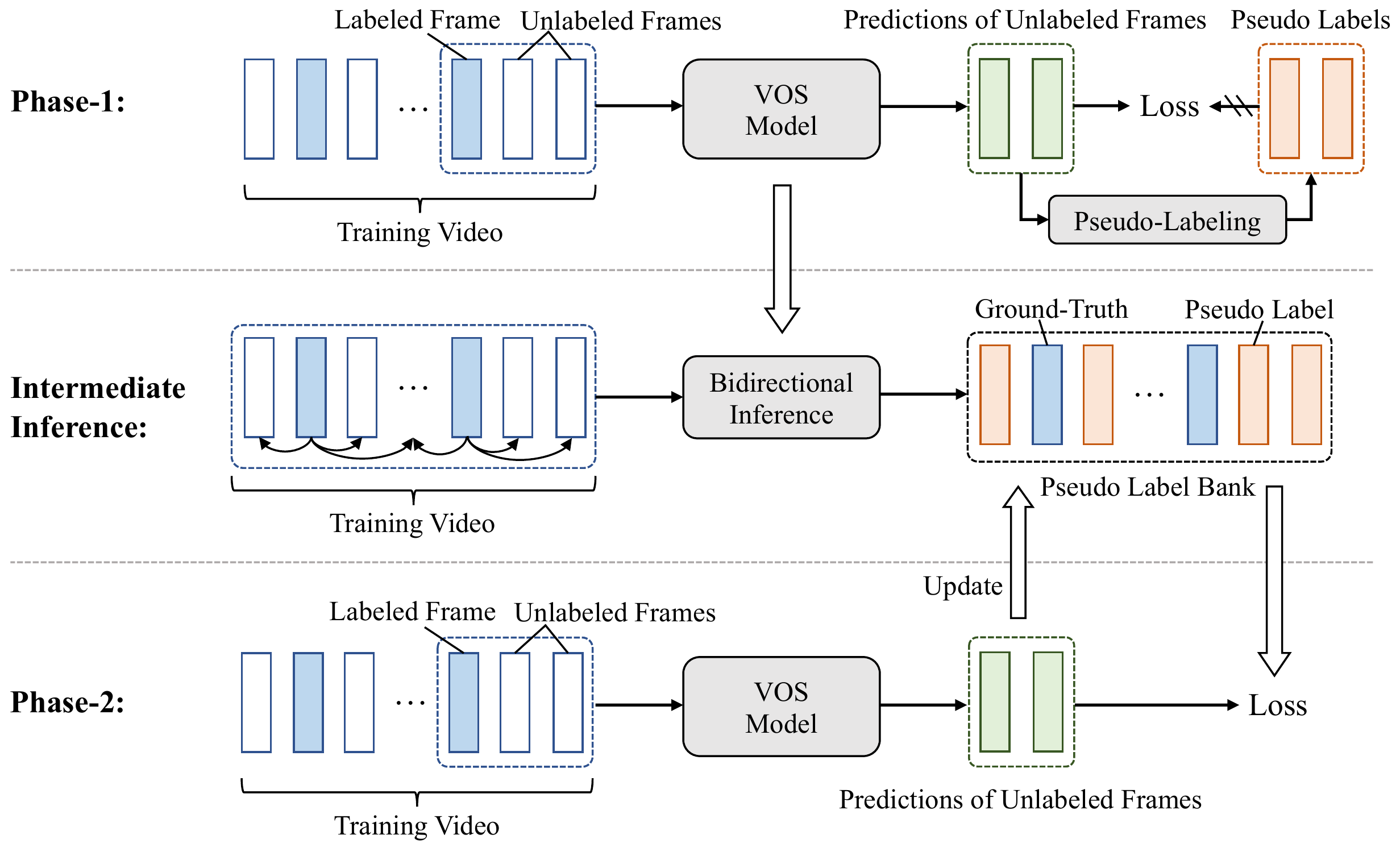}
   \caption{Overview of our methodology. In phase-1 (top), we train a VOS model (\textit{i.e.} STCN) which takes a triplet of frames as input on a two-shot VOS dataset in a semi-supervised manner. We constrain the reference (first) frame to be a labeled frame to ease the learning. 
   The remaining frames can be either labeled or unlabeled. 
   Then we perform an intermediate inference (middle) to generate pseudo labels for unlabeled frames by the VOS model trained in phase-1, and construct a pseudo-label bank to store the pseudo labels in addition to the ground-truth.
   In phase-2 (bottom), we re-train a VOS model—which could be most models—on the combination of labeled and pseudo-labeled data without any restrictions on the first frame. The pseudo-label bank is dynamically updated once more reliable pseudo labels are identified during phase-2 training.}
   \label{fig:overview}
\end{figure*}
We first revisit the preliminary of VOS in Section~\ref{preliminary}. Then we formulate the problem of two-shot VOS and show an overview of our method in Section~\ref{sec:formulation}. Next, the details of training a two-shot VOS model are presented in Section~\ref{sec:phase_one} and \ref{phase2}. At last, we show our methodology can be generalized to a majority of VOS models in Section~\ref{sec:gene}.

\subsection{Preliminary}
\label{preliminary}
Previous works train VOS models on densely annotated videos. Given the annotation of the first frame, the training objective is to maximize the mask prediction of the target object from the second frame to the last frame. For instance, STM~\cite{oh2019video} and STCN~\cite{cheng2021rethinking} take a triplet of frames as input; RDE-VOS~\cite{li2022recurrent} and XMem~\cite{cheng2022xmem} propose to model longer video sequences containing 5 and 8 frames, respectively. Random frame skipping, which randomly skips frames during the sampling, is a widely-used data augmentation to improve the generalization. In general, the maximum number of frames to skip gradually increases from 0 to $K$ as training progresses.

In our setting, we could only access two labeled frames per video. To reduce error propagation caused by unreliable pseudo labels, we adopt STCN~\cite{cheng2021rethinking} as our base model in phase-1 training since it merely needs a triplet of frames as input. Nevertheless, we can train any VOS models in phase-2, which will be described in Section~\ref{sec:gene}. Now we briefly revisit STCN. Given a training video, STCN first samples a triplet of frames as input. Then it predicts the mask of the second frame according to the ground-truth of the first frame, and the mask of the third frame based on the prediction of the previous frame in addition to the ground-truth of the first frame. The objective function of STCN is a standard segmentation loss, which is applied to each of the two predictions.

\subsection{Problem formulation and overview}
\label{sec:formulation}
\noindent\textbf{Problem formulation.} Given a VOS dataset $\mathcal{D}$, for each training video $\mathcal{V} = [\boldsymbol{V}_1, ...,\boldsymbol{V}_T] \in \mathcal{D}$ containing $T~(T\gg2)$ frames with the associated ground-truth $\mathcal{Y} = [\boldsymbol{Y}_1,...,\boldsymbol{Y}_T]$, we randomly sample two frames as the labeled data, while the remaining ones are served as the unlabeled data. The objective is to train a VOS model by using both labeled and unlabeled data.

\noindent\textbf{Overview.} \cref{fig:overview} shows an overview of our two-shot video object segmentation (VOS). First, we train a VOS model in a semi-supervised manner, with the reference frame always being a labeled one, which is referred to as \textit{phase-1} training. Then, we perform an \textit{intermediate inference} to generate pseudo labels for unlabeled frames by the VOS model trained in phase-1. The generated pseudo labels are stored in a pseudo-label bank for the convenience of accessing. 
At last, we re-train a VOS model on both labeled frames and pseudo-labeled frames without any restrictions on the reference frame. We term this stage as \textit{phase-2} training. It is worth noting that the pseudo-label bank is dynamically updated once more reliable pseudo labels are yielded in phase-2 training.

\subsection{Phase-1 training}
\label{sec:phase_one}
We adopt STCN~\cite{cheng2021rethinking} as our base model, which takes a triplet of frames as input. Nevertheless, in our setting, each training video only contains two labeled frames, which is insufficient to be served as the input of STCN in a fully supervised manner. To tackle this problem, we adopt semi-supervised learning, which generates pseudo-labeled frames together with the labeled ones to enable triplet construction. Since STCN requires the annotation of the reference (or first) frame to segment the object of interest that appeared in subsequent frames, we always use a labeled frame as the reference frame to alleviate the error propagation in the phase-1 training. The last two frames, however, can be either labeled or unlabeled. In our implementation, the last two frames have a $0.5$ probability of being both unlabeled, and a $0.5$ probability of having one frame be labeled. The training of two-shot VOS is identical to that of full-set VOS, except that our training triplet is composed of labeled frames with ground-truth and unlabeled frames with pseudo labels. Concretely, given a randomly sampled triplet where the last two frames are composed of $N_1$ labeled frames and $N_2$ unlabeled frames ($N_1=1$, $N_2=1$ or $N_1=0$, $N_2=2$), the overall loss $\mathcal{L}$ is the sum of the supervised loss $\mathcal{L}_S$ and the unsupervised loss $\mathcal{L}_U$ effected on labeled and unlabeled frames, respectively. $\mathcal{L}_S$ is a standard segmentation loss, which can be formulated as: 
\begin{equation}
    \mathcal{L}_S = \frac{1}{HWN_1}\sum_{n=1}^{N_1}\sum_{i=1}^{H}\sum_{j=1}^{W}\mathcal{H}(\boldsymbol{Y}_n^{(i,j)}, \boldsymbol{P}_n^{(i,j)}),
\label{eq:loss_l}
\end{equation}
where $H$ and $W$ represent the height and the width of the input, $\mathcal{H}(\cdot,\cdot)$ denotes the cross-entropy function, $\boldsymbol{P}_n^{(i,j)}$ is the prediction at pixel $(i,j)$ in the $n$-th labeled frame, and $\boldsymbol{Y}_n^{(i,j)}$ denotes the corresponding ground-truth. 

The unsupervised loss $\mathcal{L}_U$ is a variant of $\mathcal{L}_S$, which is defined as follows:
\begin{equation}
    \mathcal{L}_U =     \frac{1}{HWN_2}\sum_{n=1}^{N_2}\sum_{i=1}^{H}\sum_{j=1}^{W}  \mathbbm{1}_{[\max(\boldsymbol{P}_n^{(i,j)}) \geq \tau_{1}]}\mathcal{H}(\hat{\boldsymbol{Y}}_n^{(i,j)}, \boldsymbol{P}_n^{(i,j)}),
\end{equation}
where $\mathbbm{1}_{[\cdot]}$ is the indicator function to filter out the predictions whose maximal confidences are lower than the pre-defined threshold $\tau_{1}$, $\boldsymbol{P}_n^{(i,j)}$ is the prediction at pixel $(i,j)$ in the $n$-th unlabeled frame, and $\hat{\boldsymbol{Y}}_n^{(i,j)} = \mathrm{argmax}(\boldsymbol{P}_n^{(i,j)})$ represents the corresponding one-hot pseudo label. By default, we set $\tau_{1} = 0.9$ to guarantee the reliability of the yielded pseudo labels. 

As training progresses, an increasing number of high-quality pseudo-labeled samples are generated, injecting implicit knowledge included in unlabeled data into the model. In addition, we also randomly skip frames during sampling as described in Section~\ref{preliminary}.

\subsection{Phase-2 training}
\label{phase2}
\noindent\textbf{Discussion.} In phase-1 training, we constrain the reference (or first) frame to be a labeled frame since the predictions of the subsequent frames significantly rely on the mask of the reference frame. Adopting an unlabeled frame with pseudo-labeled mask as the reference frame aggravates error propagation in the early training. To make full use of the unlabeled data, we present phase-2 training, which lifts the restriction placed on the reference frame, allowing it to be either a labeled or pseudo-labeled frame. The underlying idea behind phase-2 training is to generate pseudo labels for all unlabeled frames using the decent VOS model trained in phase-1. After then, the pseudo-labeled data is stored in a pseudo-label bank, providing efficient access when constructing a training triplet where the reference frame is selected as a pseudo-labeled one.

\begin{figure}[t]
  \centering
  \includegraphics[width=0.7\columnwidth]{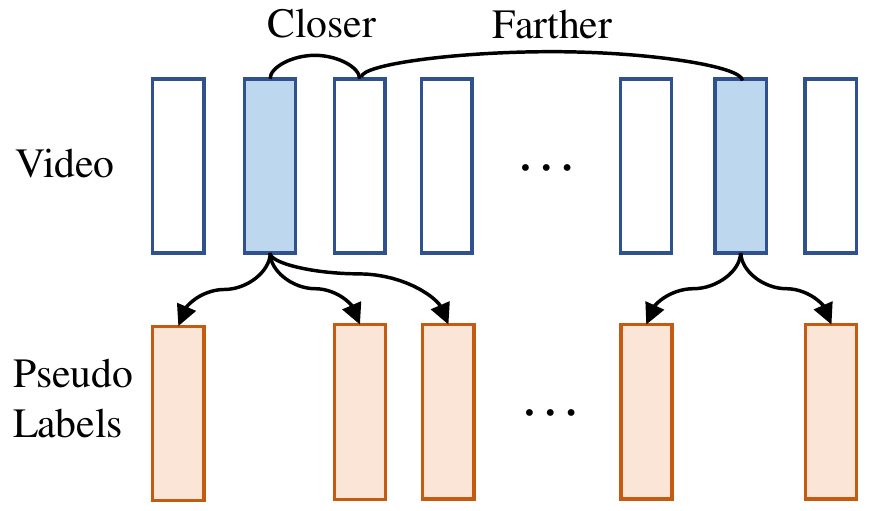}
  \caption{Illustration of bidirectional inference. Two reference frames are denoted by blue rectangles. A pre-trained VOS model infers unlabeled frames from the inference frame to the end frame and, in a reverse manner, from the inference frame to the beginning frame. We pick the prediction inferred by the labeled frame that is closest to the unlabeled frame. } \label{fig:bd}
\end{figure}

\noindent\textbf{Intermediate inference and pseudo-label bank.} We perform an intermediate inference before initiating phase-2 training. The inference of a VOS model requires the annotation of the reference (or first) frame. Nevertheless, only two labeled frames per video are available in our scenario. To generate the pseudo label per frame, inspired by bidirectional prediction and labelling~\cite{lee2022iteratively,miao2021vspw}, we introduce a bidirectional inference strategy as shown in Figure~\ref{fig:bd}. Specifically, for each of the two labeled frames, the VOS model trained in phase-1 takes it as the reference frame to infer the predictions for the unlabeled frames from the inference frame to the end frame and, in a reverse manner, from the inference frame to the beginning frame. After that, each unlabeled frame has two predictions associated with it, and we pick the prediction inferred by the labeled frame that is closest to this unlabeled frame. We maintain a pseudo-label bank to store pseudo labels associated with unlabeled frames.

\noindent\textbf{Training.} The training process of phase-2 is identical to that of phase-1, except that the reference (or first) frame can be either a labeled frame or an unlabeled frame with a pseudo label from the pseudo label bank attached to it.

\noindent\textbf{Update pseudo-label bank.}
As training progresses, predictions become more accurate, resulting in more reliable pseudo labels. Therefore, to further facilitate phase-2 training, we propose to dynamically update the pseudo-label bank as needed. Concretely, at each iteration, given the prediction $\boldsymbol{P}$ of an unlabeled frame, we use $\boldsymbol{P}^{(i,j)}$ to denote the prediction at pixel $(i,j)$. Once the prediction $\boldsymbol{P}^{(i,j)}$ meets the condition that $\mathrm{max}(\boldsymbol{P}^{(i,j)}) \geq \tau_{2}$, where $\tau_{2}$ denotes a pre-defined threshold, the corresponding pseudo label in pseudo label bank is updated by $\hat{\boldsymbol{Y}}^{(i,j)} = \mathrm{argmax}(\boldsymbol{P}^{(i,j)})$. We set $\tau_2 = 0.99$ by default.

\subsection{Generalization capability}
\label{sec:gene}
Thanks to the proposed pseudo-label bank and phase-2 training, our two-shot training paradigm can be applied to a majority of VOS models regardless of their architectures and requirements on the input. To generalize to other models, we adopt a STCN model trained in phase-1 to construct a pseudo-label bank. After that, various VOS models can utilize the universal training paradigm presented in phase-2 to enable two-shot VOS learning. Experimentally, we also apply our methodology to RDE-VOS~\cite{li2022recurrent} and XMem~\cite{cheng2022xmem} besides STCN~\cite{cheng2021rethinking} to show the generalization capability.

\begin{table*}[t]
\centering
\footnotesize
\begin{tabular}{l c  p{25pt}p{25pt}p{25pt}p{25pt}p{25pt}  p{25pt}p{25pt}p{25pt}p{25pt}p{25pt}}
\toprule
 \multirow{2}[3]{*}{Method} & \multirow{2}[3]{*}{\makecell[c]{Labeled \\ data}}& \multicolumn{5}{c}{\textbf{YouTube-VOS 2018}} & \multicolumn{5}{c}{\textbf{YouTube-VOS 2019}}\\
\cmidrule(l){3-7} \cmidrule(l){8-12}
 & & $\mathcal{G}$ & $\mathcal{J}_{S}$ & $\mathcal{F}_{S}$ & $\mathcal{J}_{U}$ & $\mathcal{F}_{U}$ & $\mathcal{G}$ & $\mathcal{J}_{S}$ & $\mathcal{F}_{S}$ & $\mathcal{J}_{U}$ & $\mathcal{F}_{U}$ \\
\midrule
STM~\cite{oh2019video}              & $100\%$  & 79.4 & 79.7 & 84.2 & 72.8 & 80.9   & - & - & - & -  & - \\ 
MiVOS~\cite{cheng2021modular}       & $100\%$  & 80.4 & 80.0 & 84.6 & 74.8 & 82.4   & 80.3 & 79.3 & 83.7 & 75.3  & 82.8 \\ 
CFBI~\cite{yang2020collaborative}   & $100\%$  & 81.4 & 81.1 & 85.8 & 75.3 & 83.4   & 81.0 & 80.6 & 85.1 & 75.2  & 83.0 \\ 
RDE-VOS~\cite{li2022recurrent}     	& $100\%$ & - & - & - & - & -     & 81.9 & 81.1 & 85.5 & 76.2 & 84.8 \\ 
HMMN~\cite{seong2021hierarchical}   & $100\%$  & 82.6 & 82.1 & 87.0 & 76.8 & 84.6   & 82.5 & 81.7 & 86.1 & 77.3 & 85.0 \\ 
JOINT~\cite{mao2021joint}           & $100\%$  & 83.1 & 81.5 & 85.9 & 78.7 & 86.5   & 82.7 & 81.1 & 85.4 & 78.2 & 85.9 \\
STCN~\cite{cheng2021rethinking}     & $100\%$  & 83.0 & 81.9 & 86.5 & 77.9 & 85.7   & 82.7 & 81.1 & 85.4 & 78.2 & 85.9 \\  
R50-AOT-L~\cite{yang2021associating}& $100\%$  & 84.1 & 83.7 & 88.5 & 78.1 & 86.1   & 84.1 & 83.5 & 88.1 & 78.4 & 86.3 \\
XMem~\cite{cheng2022xmem}           & $100\%$  & 85.7 & 84.6 & 89.3 & 80.2 & 88.7  & 85.5 & 84.3 & 88.6 & 80.3  & 88.6  \\ 
\midrule
STCN$^{*}$~\cite{cheng2021rethinking}    & $100\%$  & 83.0 & 82.0 & 86.5 & 77.8 & 85.8     & 82.7 & 81.2 & 85.4 & 78.2 & 86.0 \\  
2-shot STCN$^{*}$~\cite{cheng2021rethinking}    & $\textbf{7.3}\%$ & 80.8 & 79.5 & 83.9 & 75.9 & 84.0     & 80.6 & 79.5 & 83.8 & 75.6 & 83.4 \\  
2-shot STCN w/ Ours & $\textbf{7.3}\%$ & 82.9\textcolor{red}{$_{+2.1}$} & 81.6\textcolor{red}{$_{+2.1}$} & 86.3\textcolor{red}{$_{+2.4}$} & 77.7\textcolor{red}{$_{+1.8}$} & 86.0\textcolor{red}{$_{+2.0}$}     & 82.7\textcolor{red}{$_{+2.1}$} & 80.9\textcolor{red}{$_{+1.4}$} & 85.1\textcolor{red}{$_{+1.3}$} & 78.3\textcolor{red}{$_{+2.7}$} & 86.6\textcolor{red}{$_{+3.2}$} \\
\midrule
RDE-VOS$^{*}$~\cite{li2022recurrent}    & $100\%$  & - & - & - & - & -     & 82.1 & 81.3 & 85.7 & 76.2 & 85.0 \\  
2-shot RDE-VOS$^{*}$~\cite{li2022recurrent}    & $\textbf{7.3}\%$ & - & - & - & - & -     & 78.4 & 77.2 & 81.3 & 73.4 & 81.7 \\  
2-shot RDE-VOS w/ Ours & $\textbf{7.3}\%$ & - & - & - & - & -     & 82.1\textcolor{red}{$_{+3.7}$} & 80.4\textcolor{red}{$_{+3.2}$}  & 84.8\textcolor{red}{$_{+3.5}$}  & 77.3\textcolor{red}{$_{+3.9}$}  & 85.8\textcolor{red}{$_{+4.1}$}   \\
\midrule
XMem$^{*}$~\cite{cheng2022xmem}      & $100\%$ & 85.5 & 84.4 & 89.1 & 80.0 & 88.3  & 85.3 & 84.0 & 88.2 & 80.4  & 88.4  \\
2-shot XMem$^{*}$~\cite{cheng2022xmem}      & $\textbf{7.3}\%$ & 79.2 & 77.5 & 81.9 & 74.5 & 82.9  & 79.1 & 77.6 & 81.5 & 74.5 &  82.7 \\
2-shot XMem w/ Ours  & $\textbf{7.3}\%$ & 84.8\textcolor{red}{$_{+5.6}$} & 83.6\textcolor{red}{$_{+6.1}$}  & 88.5\textcolor{red}{$_{+6.6}$}  & 79.2\textcolor{red}{$_{+4.7}$}  & 87.7\textcolor{red}{$_{+4.8}$}   & 84.5\textcolor{red}{$_{+5.4}$}  & 83.5\textcolor{red}{$_{5.9}$}  & 88.0\textcolor{red}{$_{+6.5}$}  & 79.1\textcolor{red}{$_{+4.6}$}   & 87.3\textcolor{red}{$_{+4.6}$}   \\ 
\bottomrule
\end{tabular}
\caption{Comparison with different methods on YouTube-VOS 2018 and 2019 validation sets. Subscripts \textit{S} and \textit{U} denote seen and unseen categories respectively. $^{*}$ denotes reproduced result by using the open-source code. By using 7.3\% labeled data (2 labeled frames per training video) of YouTube-VOS benchmark, our approach achieves comparable results in contrast to the counterpart trained on full set, and outperforms the native 2-shot counterpart by large margins.}
\label{table:youtubevos}
\end{table*}

\begin{table*}[t]
\centering
\footnotesize
\begin{tabular}{l c  p{25pt}p{25pt}p{25pt}  p{25pt}p{25pt}p{25pt}}
\toprule
\multirow{2}[3]{*}{Method} & \multirow{2}[3]{*}{\makecell[c]{Labeled \\ data}} & \multicolumn{3}{c}{\textbf{DAVIS 2016}} & \multicolumn{3}{c}{\textbf{DAVIS 2017}}\\
\cmidrule(l){3-5} \cmidrule(l){6-8}
 & & $\mathcal{J} \& \mathcal{F}$ & $\mathcal{J}$ & $\mathcal{F}$ & $\mathcal{J} \& \mathcal{F}$ & $\mathcal{J}$ & $\mathcal{F}$\\
\midrule
STM~\cite{oh2019video}               & $100\%$ & 89.3 & 88.7 & 89.9   & 81.8 & 78.2 & 84.3 \\ 
CFBI~\cite{yang2020collaborative}    & $100\%$ & 89.4 & 88.3 & 90.5   & 81.9 & 79.1 & 84.6 \\ 
JOINT~\cite{mao2021joint}            & $100\%$ & -    & -    & -      & 83.5 & 80.8 & 86.2 \\ 
RDE-VOS~\cite{li2022recurrent}  & $100\%$ & 91.1  & 89.7 & 92.5    & 84.2 & 80.8 & 87.5 \\ 
MiVOS~\cite{cheng2021modular}        & $100\%$ & 91.0 & 89.6 & 92.4   & 84.5 & 81.7 & 87.4 \\ 
HMMN~\cite{seong2021hierarchical}    & $100\%$ & 90.8 & 89.6 & 92.0   & 84.7 & 81.9 & 87.5 \\ 
R50-AOT-L~\cite{yang2021associating} & $100\%$ & 91.1 & 90.1 & 92.1   & 84.9 & 82.3 & 87.5 \\ 
STCN~\cite{cheng2021rethinking}      & $100\%$ & 91.6 & 90.8 & 92.5   & 85.4 & 82.2 & 88.6 \\ 
XMem~\cite{cheng2022xmem}     & $100\%$ & 91.5 & 90.4 & 92.7  & 86.2 & 82.9 & 89.5   \\ 
\midrule
STCN$^{*}$~\cite{cheng2021rethinking}    & $100\%$ & 91.5 & 90.7 & 92.3     & 85.2 & 81.9 & 88.5 \\  
2-shot STCN$^{*}$~\cite{cheng2021rethinking}    & $\textbf{2.9}\%$ & 87.9 & 87.1 & 88.7     & 81.0 & 77.7 & 84.3 \\  
2-shot STCN w/ Ours  & $\textbf{2.9}\%$ & 91.3\textcolor{red}{$_{+3.4}$} & 90.6\textcolor{red}{$_{+3.5}$} & 92.0\textcolor{red}{$_{+3.3}$}     & 85.1\textcolor{red}{$_{+4.1}$} & 81.7\textcolor{red}{$_{+4.0}$} & 88.4\textcolor{red}{$_{+4.1}$} \\
\midrule
RDE-VOS$^{*}$~\cite{li2022recurrent}   & $100\%$ & 91.0 & 89.5 & 92.4     & 84.2 & 80.7 & 87.7 \\  
2-shot RDE-VOS$^{*}$~\cite{li2022recurrent}   & $\textbf{2.9}\%$ & 87.6 & 86.6 & 88.8     & 79.4 & 75.6 & 83.1 \\  
2-shot RDE-VOS w/ Ours  & $\textbf{2.9}\%$ & 90.8\textcolor{red}{$_{+3.2}$} & 90.0\textcolor{red}{$_{+3.4}$} & 92.0\textcolor{red}{$_{+3.2}$}    & 83.9\textcolor{red}{$_{+4.5}$} & 80.4\textcolor{red}{$_{+4.8}$} & 87.3\textcolor{red}{$_{+4.2}$} \\
\midrule
XMem$^{*}$~\cite{cheng2022xmem}      & $100\%$ & 91.3 & 90.3 & 92.4   & 86.2 & 82.8 & 89.7  \\
2-shot XMem$^{*}$~\cite{cheng2022xmem}      & $\textbf{2.9}\%$ & 88.1 & 87.1 & 89.0   & 81.7 & 78.2 & 85.1   \\
2-shot XMem w/ Ours   & $\textbf{2.9}\%$ & 91.3\textcolor{red}{$_{+3.2}$} & 90.3\textcolor{red}{$_{+3.2}$} & 92.3\textcolor{red}{$_{+3.3}$}   & 85.6\textcolor{red}{$_{+3.9}$} & 82.1\textcolor{red}{$_{+3.9}$} & 89.1\textcolor{red}{$_{+4.0}$}    \\ 
\bottomrule
\end{tabular}
\caption{Comparisons with different methods on DAVIS 2016 and 2017 validation sets. $^{*}$ denotes reproduced result by using the open-source code. By using 2.9\% labeled data (2 labeled frames per training video) of DAVIS benchmark, our approach achieves comparable results in contrast to the full-set counterpart, and outperforms the native 2-shot counterpart by large margins.
}
\label{table:davisvos}
\end{table*}
\section{Experiments}
\label{sec:expr}
\subsection{Experimental setup}
\noindent{\textbf{Datasets.}}
We conduct experiments on widely used VOS benchmarks including DAVIS 2016/2017~\cite{perazzi2016benchmark,pont20172017} and YouTube-VOS 2018/2019~\cite{xu2018youtube}.
DAVIS 2017 is a multi-object extension of DAVIS 2016, which consists of 60 (138 objects) and 30 (59 objects) videos for training and validation respectively.
YouTube-VOS is a larger-scale multi-object dataset with 3471 videos from 65 categories for training. These training videos are annotated every five frames. There are 474 and 507 videos in the 2018 and 2019 validation splits respectively. 
In our two-shot setting, we randomly select two labeled frames per video as labeled data while the remaining ones are served as unlabeled data.
Compared to full set, we only use $7.3\%$ and $2.9\%$ labeled data for YouTube-VOS and DAVIS, respectively.

\noindent{\textbf{Evaluation metric.}}
Following common practice
\cite{oh2019video,cheng2021rethinking,cheng2022xmem},
for the DAVIS datasets, we adopt the standard metrics: region similarity $\mathcal{J}$, contour accuracy $\mathcal{F}$ and their average $\mathcal{J} \& \mathcal{F}$. For the YouTube-VOS datasets, we report $\mathcal{J}$ and $\mathcal{F}$ of the seen and unseen categories, and their averaged score $\mathcal{G}$.

\noindent{\textbf{Implementation details.}}
We implement our method with PyTorch~\cite{paszke2017automatic}.
For phase-1 training, we adopt the STCN~\cite{cheng2021rethinking} pre-trained on static image datasets~\cite{wang2017learning,shi2015hierarchical,zeng2019towards} with synthetic deformations. 
The parameter $K$ in random frame skipping is gradually increased from 5 to 25 with a curriculum learning schedule. The threshold $\tau_{1}$ is set to $0.9$.
The training paradigm of two-shot VOS can be seamlessly applied to various VOS models in phase-2 training. We explore STCN~\cite{cheng2021rethinking}, RDE-VOS~\cite{li2022recurrent} and XMem~\cite{cheng2022xmem}, respectively. The threshold $\tau_{2}$ is set to $0.99$.

\begin{table}[t]
  \centering
  \footnotesize
  \begin{tabular}{c lcccc}
    \toprule
    \multirow{2}[3]{*}{Components} & \multicolumn{5}{c}{YouTube-VOS 2019} \\
    \cmidrule(l){2-6} 
     & $\mathcal{G}$ & $\mathcal{J}_{S}$ & $\mathcal{F}_{S}$ & $\mathcal{J}_{U}$ & $\mathcal{F}_{U}$ \\
    \midrule
    Baseline     & 80.6 & 79.5 & 83.8 & 75.7 & 83.4  \\
   	+phase-1  & 81.6\textcolor{red}{$_{+1.0}$} & 79.3 & 83.5 & 77.7 & 86.0  \\
    +phase-2 & 82.7\textcolor{red}{$_{+1.1}$} & 80.9 & 85.1 & 78.3 & 86.6   \\
    \bottomrule
  \end{tabular}
  \caption{Ablation study on the effectiveness of each phase. The naive 2-shot STCN is adopted as the baseline.}
  \label{tab:phase_ablation}
\end{table}
\begin{figure}[t]
  \centering
   \includegraphics[width=0.90\columnwidth]{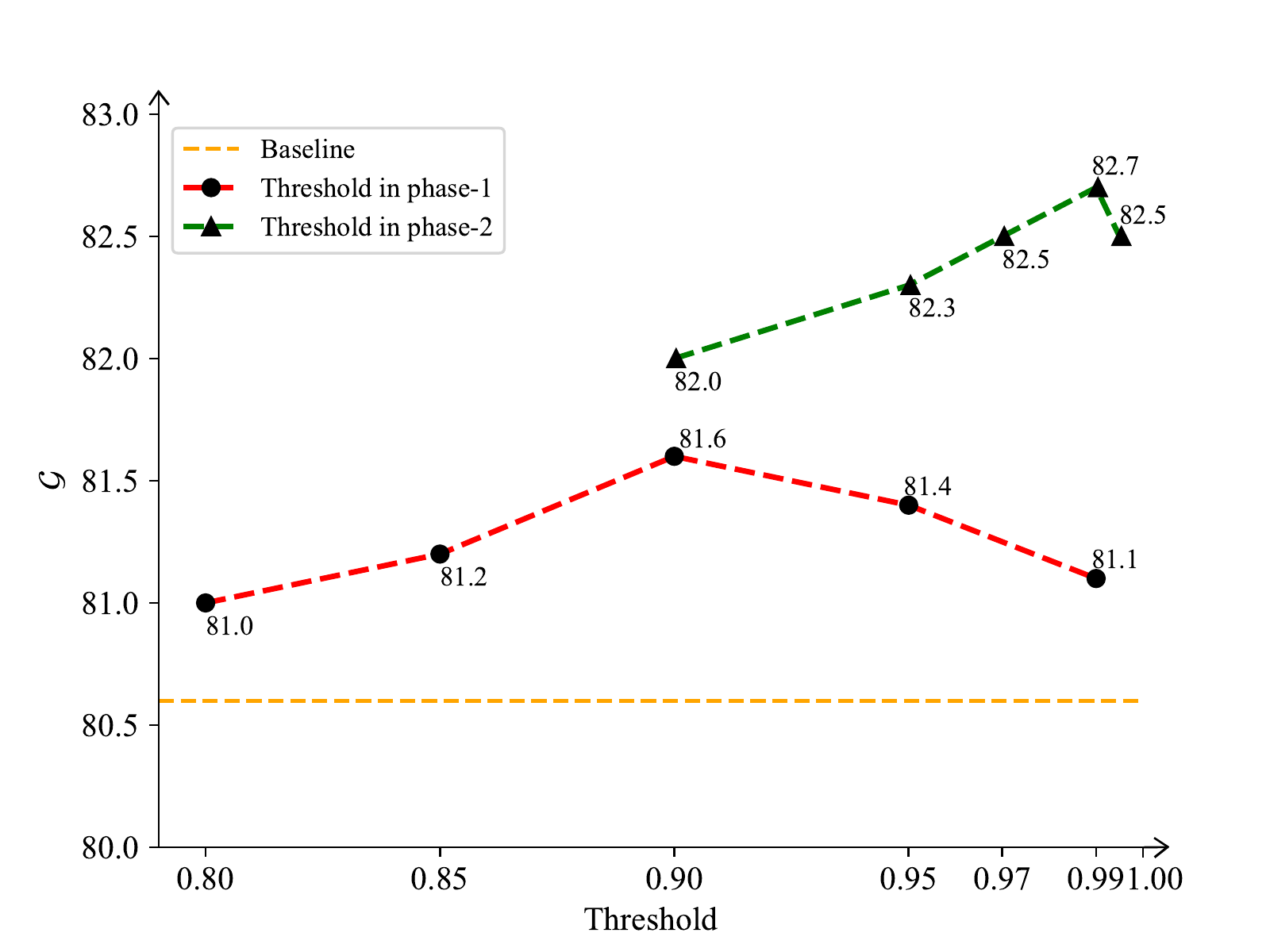}
   \caption{Study on hyper-parameters $\tau_1$ and $\tau_2$, which controls pseudo-labeling in phase-1 and -2, respectively. We adopt a higher threshold in phase-2 training since the predictions in phase-2 are more accurate than that in phase-1. By default, we set $\tau_1=0.9$ and $\tau_1=0.99$. 
   }
   \label{fig:thres}
\end{figure}

\subsection{Main results}
We apply our two-shot VOS to STCN~\cite{cheng2021rethinking}, RDE-VOS~\cite{li2022recurrent}, and XMem~\cite{cheng2022xmem}, and compare the results with 1): their counterparts trained on full sets; (2) their counterparts trained on two-shot datasets without using unlabeled data; 
(3) other strong baselines trained on full sets.
When training a naive 2-shot model in a fully supervised manner, we repeatedly sample the labeled frames to meet the input requirement of that model. We report the results on YouTube-VOS and DAVIS validation sets in \cref{table:youtubevos} and \cref{table:davisvos}, respectively. From the Tables, we could draw two conclusions: (1) Two labeled frames per video are almost sufficient for training a pleasant VOS model—even the unlabeled data are unused. For example, 2-shot STCN already achieves 80.8\% score on YouTube-VOS 2018 benchmark, which is only -2.2\% lower than the full-set STCN achieving 83.0\% score. (2) By using 7.3\% and 2.9\% labeled data of YouTube-VOS and DAVIS benchmarks, our approach achieves comparable results in contrast to the counterpart trained on full set, and outperforms the native 2-shot counterpart by large margins. For instance, 2-shot STCN equipped with our approach achieves 85.1\%/82.7\% on DAVIS 2017/YouTube-VOS 2019, which is +4.1\%/+2.1\% higher than the naive 2-shot STCN while -0.1\%/-0.0\% lower than the full-set STCN.

\begin{table}[t]
  \centering
  \footnotesize
  \begin{tabular}{c lcccc}
    \toprule
    \multirow{2}[3]{*}{Pseudo-labeler} & \multicolumn{5}{c}{YouTube-VOS 2019} \\
    \cmidrule(l){2-6} 
     & $\mathcal{G}$ & $\mathcal{J}_{S}$ & $\mathcal{F}_{S}$ & $\mathcal{J}_{U}$ & $\mathcal{F}_{U}$ \\
    \midrule
    -    & 80.6 & 79.5 & 83.8 & 75.7 & 83.4  \\
    STCN  & 81.2\textcolor{red}{$_{+0.6}$} & 79.2 & 83.5 & 77.2 & 84.9  \\
   	MT-STCN  & 81.6\textcolor{red}{$_{+0.4}$} & 79.3 & 83.5 & 77.7 & 86.0  \\
    \bottomrule
  \end{tabular}
  \caption{Ablation study of different pseudo-labelers in phase-1. MT-STCN: the parameters of STCN is
  updated by a Mean Teacher~\cite{tarvainen2017mean} strategy.}
  \label{tab:pseudo-labeler}
\end{table}

\begin{table}[t]
  \centering
  \footnotesize
  \begin{tabular}{c ccccc}
    \toprule
    \multirow{2}[3]{*}{$\alpha$} & \multicolumn{5}{c}{YouTube-VOS 2019} \\
    \cmidrule(l){2-6} 
     & $\mathcal{G}$ & $\mathcal{J}_{S}$ & $\mathcal{F}_{S}$ & $\mathcal{J}_{U}$ & $\mathcal{F}_{U}$ \\
    \midrule
     0.990  & 81.2 & 79.4 & 83.8 & 76.9 & 84.5  \\
    \midrule
     0.995 & 81.6 & 79.3 & 83.5 & 77.7 & 86.0  \\
    \midrule
     0.999 & 81.3 & 79.4 & 83.7 & 76.9 & 85.2  \\
    \bottomrule
  \end{tabular}
  \caption{Study of different coefficient $\alpha$ used in the MT-STCN.}
  \label{tab:alpha_ablation}
\end{table}

\subsection{Ablation study}
In this section, we validate the proposed two-shot VOS training strategy step-by-step.
All ablation studies are conducted on Youtube-VOS 2019 by applying our approach to STCN~\cite{cheng2021rethinking}. More analysis can be found in our supplementary material.

\noindent \textbf{Effects of each phase.} The results are shown in \cref{tab:phase_ablation}.  Starting from a naive 2-shot STCN (denoted as ``baseline'' afterward) which achieves 80.6$\%$ score, phase-1 training improves the score to 81.6$\%$. On top of this, phase-2 training further enhances performance to 82.7$\%$, leading to the same performance of STCN trained on fully labeled set. 

\noindent \textbf{Thresholds of pseudo-labeling.} There are two hyper-parameters $\tau_1$ and $\tau_2$ controlling pseudo-labeling in phase-1 and -2, respectively. \cref{fig:thres} displays two accuracy curves by varying $\tau_1$ and $\tau_2$. Using a higher threshold guarantees the quality of generated pseudo labels but yields less amount of pseudo data, and vice versa. We adopt a higher threshold in phase-2 training since the predictions in phase-2 are more accurate than that in phase-1. It can be seen that $\tau_1=0.9$ and $\tau_1=0.99$ yield the best result.

\noindent \textbf{Different pseudo labelers.} \cref{tab:pseudo-labeler} ablates the effects of using different pseudo-labelers in phase-1. Specifically, we propose two variants: (1) STCN model itself; (2) STCN with a mean teacher~\cite{tarvainen2017mean} strategy. The underlying idea behind Mean Teacher (MT) is that using an exponential moving average (EMA) strategy to update the parameters of the model at each iteration, which can be formulated as:  
$\theta_{t}^{'} = \alpha \theta_{t-1}^{'} + (1 - \alpha)\theta_{t}$,
where $t$ denotes the current iteration, $\theta_{t}^{'}$ and $\theta_{t}$ denote the parameters of MT-STCN and STCN respectively,
and $\alpha$ is a weight. It can be seen that using the MT-STCN model surpasses
the one without MT strategy. We further ablate $\alpha$ in \cref{tab:alpha_ablation}. We find that $\alpha = 0.995$ yields the best performance. However, we do not employ MT strategy in phase-2 since no performance improvement is observed.

\begin{table}[t]
  \centering
  \footnotesize
  \begin{tabular}{c lcccc}
    \toprule
    \multirow{2}[3]{*}{\makecell[c]{Intermediate \\ inference} } & \multicolumn{5}{c}{YouTube-VOS 2019} \\
    \cmidrule(l){2-6} 
     & $\mathcal{G}$ & $\mathcal{J}_{S}$ & $\mathcal{F}_{S}$ & $\mathcal{J}_{U}$ & $\mathcal{F}_{U}$ \\
    \midrule
    Unidirectional  & 82.1 & 80.8 & 77.3 & 77.6 & 85.2   \\
   	Bidirectional   & 82.7\textcolor{red}{$_{+0.6}$} & 80.9 & 85.1 & 78.3 & 86.6   \\
    \bottomrule
  \end{tabular}
  \caption{Comparison between unidirectional inference and bidirectional inference (default).}
  \label{tab:inter_infer}
\end{table}
\begin{table}[t]
  \centering
  \footnotesize
  \begin{tabular}{c lcccc}
    \toprule
    \multirow{2}[3]{*}{Update} & \multicolumn{5}{c}{YouTube-VOS 2019} \\
    \cmidrule(l){2-6} 
     & $\mathcal{G}$ & $\mathcal{J}_{S}$ & $\mathcal{F}_{S}$ & $\mathcal{J}_{U}$ & $\mathcal{F}_{U}$ \\
    \midrule
      & 82.2 & 80.7 & 84.9 & 77.6 & 85.5   \\
   	\checkmark & 82.7\textcolor{red}{$_{+0.5}$} & 80.9 & 85.1 & 78.3 & 86.6   \\
    \bottomrule
  \end{tabular}
  \caption{Study on pseudo-label bank update in phase-2 training.}
  \label{tab:label_bank}
\end{table}

\noindent \textbf{Bidirectional inference.} We adopt an intermediate inference to construct a pseudo-label bank to enable phase-2 training. We compare the proposed bidirectional inference with the unidirectional inference, which is typically used in most VOS models. The results are shown in \cref{tab:inter_infer}. There is a +0.6$\%$ improvement when utilizing bidirectional inference versus unidirectional inference. The reasons are that: (1) some unlabeled frames are not associated with the pseudo labels in the unidirectional inference; 
(2) the bidirectional inference alleviates the error propagation issue.

\noindent \textbf{Dynamically update the pseudo-label bank.} We verify the effectiveness of dynamically updating the pseudo-label bank during phase-2 training, by comparing \yankun{it} with a variant that freezes the pseudo-label bank once constructed. As shown in \cref{tab:label_bank}, freezing the pseudo-label bank slightly hurt the performance. As training progresses, more accurate pseudo labels are generated, thus it is optimal to update the pseudo-label bank to further promote the learning. 

\noindent \textbf{Visualization of feature space.} We randomly pick two unlabeled frames from the constructed 2-shot YouTube-VOS 2019 training set for feature space visualization. Note we could access their annotations (foreground and background) from the full set. We use PCA to visualize the feature space of naive 2-shot STCN, 2-shot STCN with our training paradigm, and full-set STCN in \cref{fig:feature_vis}. Both 2-shot STCN equipped with our methodology, and full-set STCN show more compact clusters.

\subsection{Discussion}

\noindent \textbf{How about more shots?} We conduct experiments under the $4$-shot and $6$-shot settings. We apply our approach to $4$- and $6$-shot STCN and conduct one round of phase-1 training. Two models achieve the performance of 82.0$\%$ and 82.1$\%$ on YouTube-VOS 2019, respectively. We further conduct one round of phase-2 training. Both models achieve 82.7$\%$ on YouTube-VOS 2019, which is the same as that of 2-shot STCN equipped with our method—the performance is already saturated for two-shot VOS and acquiring more labeled data may not be beneficial.

\noindent \textbf{Robustness of our approach.} To verify the robustness of our approach, we independently construct five 2-shot VOS datasets from YouTube-VOS 2019 benchmark and train a 2-shot STCN with our methodology on each set. The results are $[$82.69$\%$, 82.70$\%$, 82.72$\%$, 82.72$\%$, 82.73$\%$$]$, with an average of 82.71$\%$ and a standard deviation of 0.015$\%$, showing the robustness of our approach.

\begin{figure}[t]
  \centering
  \includegraphics[width=0.95\columnwidth]{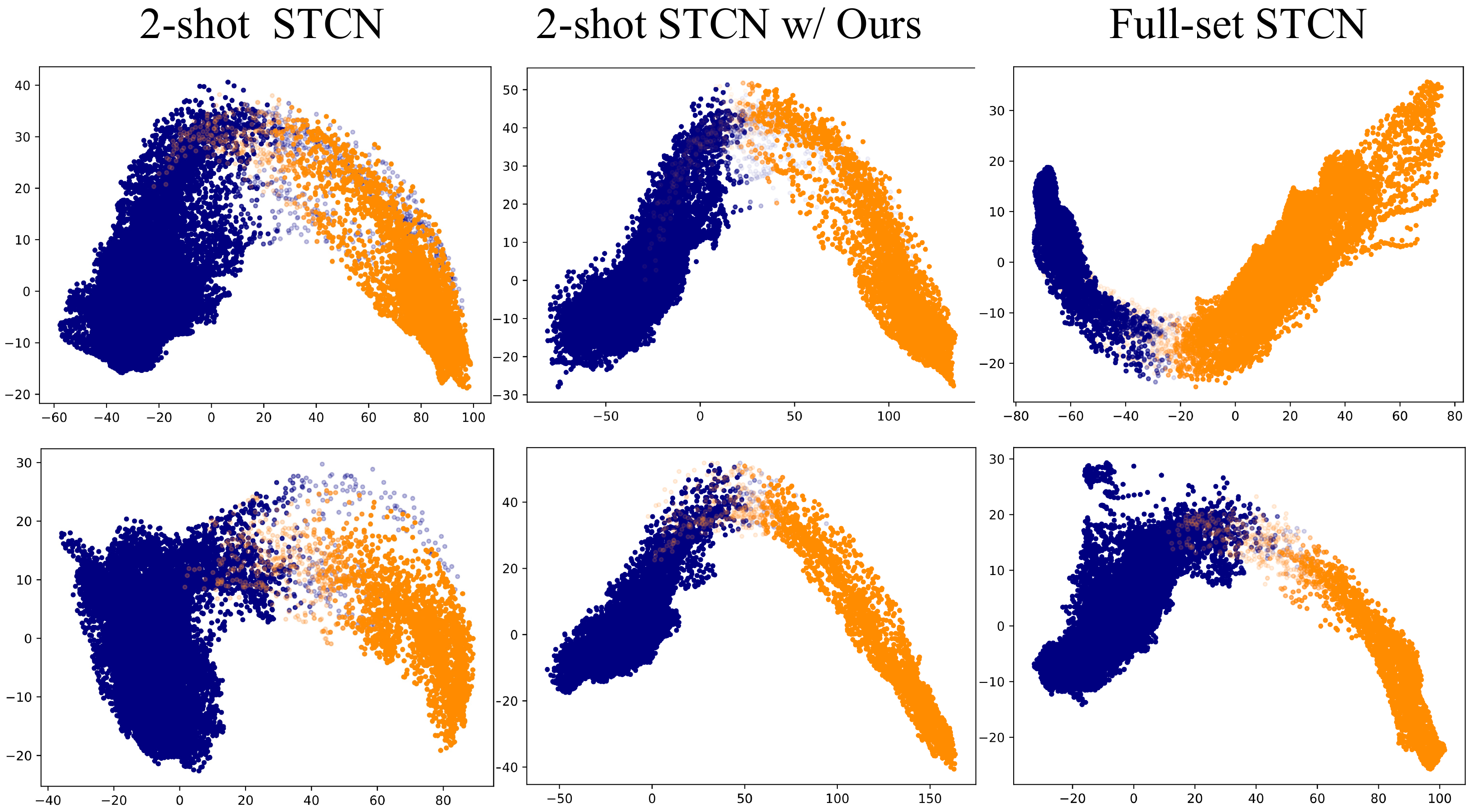}
  \caption{Feature space visualization with PCA. We compare naive 2-shot STCN, 2-shot STCN equipped with our approach, and full-set STCN. Orange:
  foreground; blue: background.}
  \label{fig:feature_vis}
\end{figure}

\section{Conclusion}
For the first time, we demonstrate the feasibility that only two labeled frames per video are almost sufficient for training a decent VOS model. On top of this, we present a simple training paradigm to resolve two-shot VOS. The underlying idea behind our approach is to exploit the wealth of information present in unlabeled data in a semi-supervised learning manner. Our approach can be applied to a majority of fully supervised VOS models, such as STCN, RDE-VOS, and XMem. By using 7.3\% and 2.9\% labeled data of YouTube-VOS and DAVIS benchmarks, our approach achieves comparable results in contrast to the counterparts trained on fully labeled set. With its simplicity and strong performance, we hope our approach can serve as a solid baseline for future research.

{\small
\bibliographystyle{ieee_fullname}
\bibliography{egbib}

\begin{thebibliography}{10}\itemsep=-1pt

\bibitem{bachman2014learning}
Philip Bachman, Ouais Alsharif, and Doina Precup.
\newblock Learning with pseudo-ensembles.
\newblock {\em Advances in neural information processing systems}, 27, 2014.

\bibitem{berthelot2019remixmatch}
David Berthelot, Nicholas Carlini, Ekin~D Cubuk, Alex Kurakin, Kihyuk Sohn, Han
  Zhang, and Colin Raffel.
\newblock Remixmatch: Semi-supervised learning with distribution alignment and
  augmentation anchoring.
\newblock {\em arXiv preprint arXiv:1911.09785}, 2019.

\bibitem{berthelot2019mixmatch}
David Berthelot, Nicholas Carlini, Ian Goodfellow, Nicolas Papernot, Avital
  Oliver, and Colin~A Raffel.
\newblock Mixmatch: A holistic approach to semi-supervised learning.
\newblock {\em Advances in neural information processing systems}, 32, 2019.

\bibitem{caelles2017one}
Sergi Caelles, Kevis-Kokitsi Maninis, Jordi Pont-Tuset, Laura Leal-Taix{\'e},
  Daniel Cremers, and Luc Van~Gool.
\newblock One-shot video object segmentation.
\newblock In {\em Proceedings of the IEEE conference on computer vision and
  pattern recognition}, pages 221--230, 2017.

\bibitem{chen2020state}
Xi Chen, Zuoxin Li, Ye Yuan, Gang Yu, Jianxin Shen, and Donglian Qi.
\newblock State-aware tracker for real-time video object segmentation.
\newblock In {\em Proceedings of the IEEE/CVF Conference on Computer Vision and
  Pattern Recognition}, pages 9384--9393, 2020.

\bibitem{chen2021semi}
Xiaokang Chen, Yuhui Yuan, Gang Zeng, and Jingdong Wang.
\newblock Semi-supervised semantic segmentation with cross pseudo supervision.
\newblock In {\em Proceedings of the IEEE/CVF Conference on Computer Vision and
  Pattern Recognition}, pages 2613--2622, 2021.

\bibitem{cheng2022xmem}
Ho~Kei Cheng and Alexander~G Schwing.
\newblock Xmem: Long-term video object segmentation with an atkinson-shiffrin
  memory model.
\newblock {\em arXiv preprint arXiv:2207.07115}, 2022.

\bibitem{cheng2021modular}
Ho~Kei Cheng, Yu-Wing Tai, and Chi-Keung Tang.
\newblock Modular interactive video object segmentation: Interaction-to-mask,
  propagation and difference-aware fusion.
\newblock In {\em Proceedings of the IEEE/CVF Conference on Computer Vision and
  Pattern Recognition}, pages 5559--5568, 2021.

\bibitem{cheng2021rethinking}
Ho~Kei Cheng, Yu-Wing Tai, and Chi-Keung Tang.
\newblock Rethinking space-time networks with improved memory coverage for
  efficient video object segmentation.
\newblock {\em Advances in Neural Information Processing Systems},
  34:11781--11794, 2021.

\bibitem{cheng2017segflow}
Jingchun Cheng, Yi-Hsuan Tsai, Shengjin Wang, and Ming-Hsuan Yang.
\newblock Segflow: Joint learning for video object segmentation and optical
  flow.
\newblock In {\em Proceedings of the IEEE international conference on computer
  vision}, pages 686--695, 2017.

\bibitem{french2019semi}
Geoff French, Samuli Laine, Timo Aila, Michal Mackiewicz, and Graham Finlayson.
\newblock Semi-supervised semantic segmentation needs strong, varied
  perturbations.
\newblock {\em arXiv preprint arXiv:1906.01916}, 2019.

\bibitem{ge2021video}
Wenbin Ge, Xiankai Lu, and Jianbing Shen.
\newblock Video object segmentation using global and instance embedding
  learning.
\newblock In {\em Proceedings of the IEEE/CVF Conference on Computer Vision and
  Pattern Recognition}, pages 16836--16845, 2021.

\bibitem{grandvalet2004semi}
Yves Grandvalet and Yoshua Bengio.
\newblock Semi-supervised learning by entropy minimization.
\newblock {\em Advances in neural information processing systems}, 17, 2004.

\bibitem{hu2021semi}
Hanzhe Hu, Fangyun Wei, Han Hu, Qiwei Ye, Jinshi Cui, and Liwei Wang.
\newblock Semi-supervised semantic segmentation via adaptive equalization
  learning.
\newblock {\em Advances in Neural Information Processing Systems},
  34:22106--22118, 2021.

\bibitem{hu2021learning}
Li Hu, Peng Zhang, Bang Zhang, Pan Pan, Yinghui Xu, and Rong Jin.
\newblock Learning position and target consistency for memory-based video
  object segmentation.
\newblock In {\em Proceedings of the IEEE/CVF Conference on Computer Vision and
  Pattern Recognition}, pages 4144--4154, 2021.

\bibitem{johnander2019generative}
Joakim Johnander, Martin Danelljan, Emil Brissman, Fahad~Shahbaz Khan, and
  Michael Felsberg.
\newblock A generative appearance model for end-to-end video object
  segmentation.
\newblock In {\em Proceedings of the IEEE/CVF Conference on Computer Vision and
  Pattern Recognition}, pages 8953--8962, 2019.

\bibitem{jolliffe2016principal}
Ian~T Jolliffe and Jorge Cadima.
\newblock Principal component analysis: a review and recent developments.
\newblock {\em Philosophical Transactions of the Royal Society A: Mathematical,
  Physical and Engineering Sciences}, 374(2065):20150202, 2016.

\bibitem{ke2020guided}
Zhanghan Ke, Di Qiu, Kaican Li, Qiong Yan, and Rynson~WH Lau.
\newblock Guided collaborative training for pixel-wise semi-supervised
  learning.
\newblock In {\em European conference on computer vision}, pages 429--445.
  Springer, 2020.

\bibitem{laine2016temporal}
Samuli Laine and Timo Aila.
\newblock Temporal ensembling for semi-supervised learning.
\newblock {\em arXiv preprint arXiv:1610.02242}, 2016.

\bibitem{lee2013pseudo}
Dong-Hyun Lee et~al.
\newblock Pseudo-label: The simple and efficient semi-supervised learning
  method for deep neural networks.
\newblock In {\em Workshop on challenges in representation learning, ICML},
  volume~3, page 896, 2013.

\bibitem{lee2022iteratively}
Youngjo Lee, Hongje Seong, and Euntai Kim.
\newblock Iteratively selecting an easy reference frame makes unsupervised
  video object segmentation easier.
\newblock In {\em AAAI}, 2022.

\bibitem{li2022recurrent}
Mingxing Li, Li Hu, Zhiwei Xiong, Bang Zhang, Pan Pan, and Dong Liu.
\newblock Recurrent dynamic embedding for video object segmentation.
\newblock In {\em Proceedings of the IEEE/CVF Conference on Computer Vision and
  Pattern Recognition}, pages 1332--1341, 2022.

\bibitem{li2018video}
Xiaoxiao Li and Chen~Change Loy.
\newblock Video object segmentation with joint re-identification and
  attention-aware mask propagation.
\newblock In {\em Proceedings of the European conference on computer vision
  (ECCV)}, pages 90--105, 2018.

\bibitem{lu2020video}
Xiankai Lu, Wenguan Wang, Martin Danelljan, Tianfei Zhou, Jianbing Shen, and
  Luc~Van Gool.
\newblock Video object segmentation with episodic graph memory networks.
\newblock In {\em European Conference on Computer Vision}, pages 661--679.
  Springer, 2020.

\bibitem{luiten2018premvos}
Jonathon Luiten, Paul Voigtlaender, and Bastian Leibe.
\newblock Premvos: Proposal-generation, refinement and merging for video object
  segmentation.
\newblock In {\em Asian Conference on Computer Vision}, pages 565--580.
  Springer, 2018.

\bibitem{maninis2018video}
K-K Maninis, Sergi Caelles, Yuhua Chen, Jordi Pont-Tuset, Laura Leal-Taix{\'e},
  Daniel Cremers, and Luc Van~Gool.
\newblock Video object segmentation without temporal information.
\newblock {\em IEEE transactions on pattern analysis and machine intelligence},
  41(6):1515--1530, 2018.

\bibitem{mao2021joint}
Yunyao Mao, Ning Wang, Wengang Zhou, and Houqiang Li.
\newblock Joint inductive and transductive learning for video object
  segmentation.
\newblock In {\em Proceedings of the IEEE/CVF International Conference on
  Computer Vision}, pages 9670--9679, 2021.

\bibitem{miao2021vspw}
Jiaxu Miao, Yunchao Wei, Yu Wu, Chen Liang, Guangrui Li, and Yi Yang.
\newblock Vspw: A large-scale dataset for video scene parsing in the wild.
\newblock In {\em CVPR}, 2021.

\bibitem{miyato2018virtual}
Takeru Miyato, Shin-ichi Maeda, Masanori Koyama, and Shin Ishii.
\newblock Virtual adversarial training: a regularization method for supervised
  and semi-supervised learning.
\newblock {\em IEEE transactions on pattern analysis and machine intelligence},
  41(8):1979--1993, 2018.

\bibitem{oh2018fast}
Seoung~Wug Oh, Joon-Young Lee, Kalyan Sunkavalli, and Seon~Joo Kim.
\newblock Fast video object segmentation by reference-guided mask propagation.
\newblock In {\em Proceedings of the IEEE conference on computer vision and
  pattern recognition}, pages 7376--7385, 2018.

\bibitem{oh2019video}
Seoung~Wug Oh, Joon-Young Lee, Ning Xu, and Seon~Joo Kim.
\newblock Video object segmentation using space-time memory networks.
\newblock In {\em Proceedings of the IEEE/CVF International Conference on
  Computer Vision}, pages 9226--9235, 2019.

\bibitem{paszke2017automatic}
Adam Paszke, Sam Gross, Soumith Chintala, Gregory Chanan, Edward Yang, Zachary
  DeVito, Zeming Lin, Alban Desmaison, Luca Antiga, and Adam Lerer.
\newblock Automatic differentiation in pytorch.
\newblock 2017.

\bibitem{perazzi2017learning}
Federico Perazzi, Anna Khoreva, Rodrigo Benenson, Bernt Schiele, and Alexander
  Sorkine-Hornung.
\newblock Learning video object segmentation from static images.
\newblock In {\em Proceedings of the IEEE conference on computer vision and
  pattern recognition}, pages 2663--2672, 2017.

\bibitem{perazzi2016benchmark}
Federico Perazzi, Jordi Pont-Tuset, Brian McWilliams, Luc Van~Gool, Markus
  Gross, and Alexander Sorkine-Hornung.
\newblock A benchmark dataset and evaluation methodology for video object
  segmentation.
\newblock In {\em Proceedings of the IEEE conference on computer vision and
  pattern recognition}, pages 724--732, 2016.

\bibitem{pont20172017}
Jordi Pont-Tuset, Federico Perazzi, Sergi Caelles, Pablo Arbel{\'a}ez, Alex
  Sorkine-Hornung, and Luc Van~Gool.
\newblock The 2017 davis challenge on video object segmentation.
\newblock {\em arXiv preprint arXiv:1704.00675}, 2017.

\bibitem{sajjadi2016regularization}
Mehdi Sajjadi, Mehran Javanmardi, and Tolga Tasdizen.
\newblock Regularization with stochastic transformations and perturbations for
  deep semi-supervised learning.
\newblock {\em Advances in neural information processing systems}, 29, 2016.

\bibitem{seong2020kernelized}
Hongje Seong, Junhyuk Hyun, and Euntai Kim.
\newblock Kernelized memory network for video object segmentation.
\newblock In {\em European Conference on Computer Vision}, pages 629--645.
  Springer, 2020.

\bibitem{seong2021hierarchical}
Hongje Seong, Seoung~Wug Oh, Joon-Young Lee, Seongwon Lee, Suhyeon Lee, and
  Euntai Kim.
\newblock Hierarchical memory matching network for video object segmentation.
\newblock In {\em Proceedings of the IEEE/CVF International Conference on
  Computer Vision}, pages 12889--12898, 2021.

\bibitem{shi2015hierarchical}
Jianping Shi, Qiong Yan, Li Xu, and Jiaya Jia.
\newblock Hierarchical image saliency detection on extended cssd.
\newblock {\em IEEE transactions on pattern analysis and machine intelligence},
  38(4):717--729, 2015.

\bibitem{sohn2020fixmatch}
Kihyuk Sohn, David Berthelot, Nicholas Carlini, Zizhao Zhang, Han Zhang,
  Colin~A Raffel, Ekin~Dogus Cubuk, Alexey Kurakin, and Chun-Liang Li.
\newblock Fixmatch: Simplifying semi-supervised learning with consistency and
  confidence.
\newblock {\em Advances in neural information processing systems}, 33:596--608,
  2020.

\bibitem{sohn2020simple}
Kihyuk Sohn, Zizhao Zhang, Chun-Liang Li, Han Zhang, Chen-Yu Lee, and Tomas
  Pfister.
\newblock A simple semi-supervised learning framework for object detection.
\newblock {\em arXiv preprint arXiv:2005.04757}, 2020.

\bibitem{tarvainen2017mean}
Antti Tarvainen and Harri Valpola.
\newblock Mean teachers are better role models: Weight-averaged consistency
  targets improve semi-supervised deep learning results.
\newblock {\em Advances in neural information processing systems}, 30, 2017.

\bibitem{voigtlaender2017online}
Paul Voigtlaender and Bastian Leibe.
\newblock Online adaptation of convolutional neural networks for video object
  segmentation.
\newblock {\em arXiv preprint arXiv:1706.09364}, 2017.

\bibitem{wang2021swiftnet}
Haochen Wang, Xiaolong Jiang, Haibing Ren, Yao Hu, and Song Bai.
\newblock Swiftnet: Real-time video object segmentation.
\newblock In {\em Proceedings of the IEEE/CVF Conference on Computer Vision and
  Pattern Recognition}, pages 1296--1305, 2021.

\bibitem{wang2017learning}
Lijun Wang, Huchuan Lu, Yifan Wang, Mengyang Feng, Dong Wang, Baocai Yin, and
  Xiang Ruan.
\newblock Learning to detect salient objects with image-level supervision.
\newblock In {\em Proceedings of the IEEE conference on computer vision and
  pattern recognition}, pages 136--145, 2017.

\bibitem{xiao2018monet}
Huaxin Xiao, Jiashi Feng, Guosheng Lin, Yu Liu, and Maojun Zhang.
\newblock Monet: Deep motion exploitation for video object segmentation.
\newblock In {\em Proceedings of the IEEE Conference on Computer Vision and
  Pattern Recognition}, pages 1140--1148, 2018.

\bibitem{xie2021efficient}
Haozhe Xie, Hongxun Yao, Shangchen Zhou, Shengping Zhang, and Wenxiu Sun.
\newblock Efficient regional memory network for video object segmentation.
\newblock In {\em Proceedings of the IEEE/CVF Conference on Computer Vision and
  Pattern Recognition}, pages 1286--1295, 2021.

\bibitem{xie2020unsupervised}
Qizhe Xie, Zihang Dai, Eduard Hovy, Thang Luong, and Quoc Le.
\newblock Unsupervised data augmentation for consistency training.
\newblock {\em Advances in Neural Information Processing Systems},
  33:6256--6268, 2020.

\bibitem{xie2020self}
Qizhe Xie, Minh-Thang Luong, Eduard Hovy, and Quoc~V Le.
\newblock Self-training with noisy student improves imagenet classification.
\newblock In {\em Proceedings of the IEEE/CVF conference on computer vision and
  pattern recognition}, pages 10687--10698, 2020.

\bibitem{xu2021end}
Mengde Xu, Zheng Zhang, Han Hu, Jianfeng Wang, Lijuan Wang, Fangyun Wei, Xiang
  Bai, and Zicheng Liu.
\newblock End-to-end semi-supervised object detection with soft teacher.
\newblock In {\em Proceedings of the IEEE/CVF International Conference on
  Computer Vision}, pages 3060--3069, 2021.

\bibitem{xu2018youtube}
Ning Xu, Linjie Yang, Yuchen Fan, Dingcheng Yue, Yuchen Liang, Jianchao Yang,
  and Thomas Huang.
\newblock Youtube-vos: A large-scale video object segmentation benchmark.
\newblock {\em arXiv preprint arXiv:1809.03327}, 2018.

\bibitem{xu2022cross}
Yinghao Xu, Fangyun Wei, Xiao Sun, Ceyuan Yang, Yujun Shen, Bo Dai, Bolei Zhou,
  and Stephen Lin.
\newblock Cross-model pseudo-labeling for semi-supervised action recognition.
\newblock In {\em Proceedings of the IEEE/CVF Conference on Computer Vision and
  Pattern Recognition}, pages 2959--2968, 2022.

\bibitem{yang2020collaborative}
Zongxin Yang, Yunchao Wei, and Yi Yang.
\newblock Collaborative video object segmentation by foreground-background
  integration.
\newblock In {\em European Conference on Computer Vision}, pages 332--348.
  Springer, 2020.

\bibitem{yang2021associating}
Zongxin Yang, Yunchao Wei, and Yi Yang.
\newblock Associating objects with transformers for video object segmentation.
\newblock {\em Advances in Neural Information Processing Systems},
  34:2491--2502, 2021.

\bibitem{zeng2019towards}
Yi Zeng, Pingping Zhang, Jianming Zhang, Zhe Lin, and Huchuan Lu.
\newblock Towards high-resolution salient object detection.
\newblock In {\em Proceedings of the IEEE/CVF International Conference on
  Computer Vision}, pages 7234--7243, 2019.

\bibitem{zhang2020transductive}
Yizhuo Zhang, Zhirong Wu, Houwen Peng, and Stephen Lin.
\newblock A transductive approach for video object segmentation.
\newblock In {\em Proceedings of the IEEE/CVF Conference on Computer Vision and
  Pattern Recognition}, pages 6949--6958, 2020.

\bibitem{zoph2020rethinking}
Barret Zoph, Golnaz Ghiasi, Tsung-Yi Lin, Yin Cui, Hanxiao Liu, Ekin~Dogus
  Cubuk, and Quoc Le.
\newblock Rethinking pre-training and self-training.
\newblock {\em Advances in neural information processing systems},
  33:3833--3845, 2020.

\bibitem{zou2020pseudoseg}
Yuliang Zou, Zizhao Zhang, Han Zhang, Chun-Liang Li, Xiao Bian, Jia-Bin Huang,
  and Tomas Pfister.
\newblock Pseudoseg: Designing pseudo labels for semantic segmentation.
\newblock {\em arXiv preprint arXiv:2010.09713}, 2020.

\end{thebibliography}
}

\appendix

\section{One-shot VOS}
One-shot VOS is meaningless as it degrades VOS to the task of image-level segmentation. However, we can still train a VOS model by augmenting a single labeled frame into a video clip. Experimentally, we train a naive one-shot STCN, which achieves 72.9$\%$ score on YouTube-VOS 2019 and performs -7.7$\%$ lower than its native two-shot counterpart. In addition, we randomly sample half of the training videos from YouTube-VOS 2019 dataset and train another native two-shot STCN model on this subset. This two-shot STCN model, which uses the same amount of labeled data as one-shot STCN, achieves 79.8$\%$ score, outperforming the one-shot counterpart by +6.9\%. This study verifies that it is non-trivial to directly apply our approach to the one-shot scenario because, if no adjustments are adopted to the existing VOS models, the learning of VOS necessitates at least two labeled frames. We leave the one-shot VOS for future research.

\section{Visualization}
\label{sec:supple}
\noindent \textbf{Visualization of feature space.} 
We randomly pick five unlabeled frames from the constructed 2-shot YouTube-VOS 2019~\cite{xu2018youtube} training set for feature space visualization. Note we could access their annotations (foreground and background) from the fully labeled set. We adopt PCA~\cite{jolliffe2016principal} to reduce the dimension of the pixel-wise features from 256 to 2. We visualize the feature space of naive 2-shot STCN, 2-shot STCN with our training paradigm, and full-set STCN in \cref{fig:feature_vis}. Both 2-shot STCN equipped with our methodology, and full-set STCN show more compact clusters.

\begin{figure*}[t]
  \centering
  \includegraphics[width=1.0\textwidth]{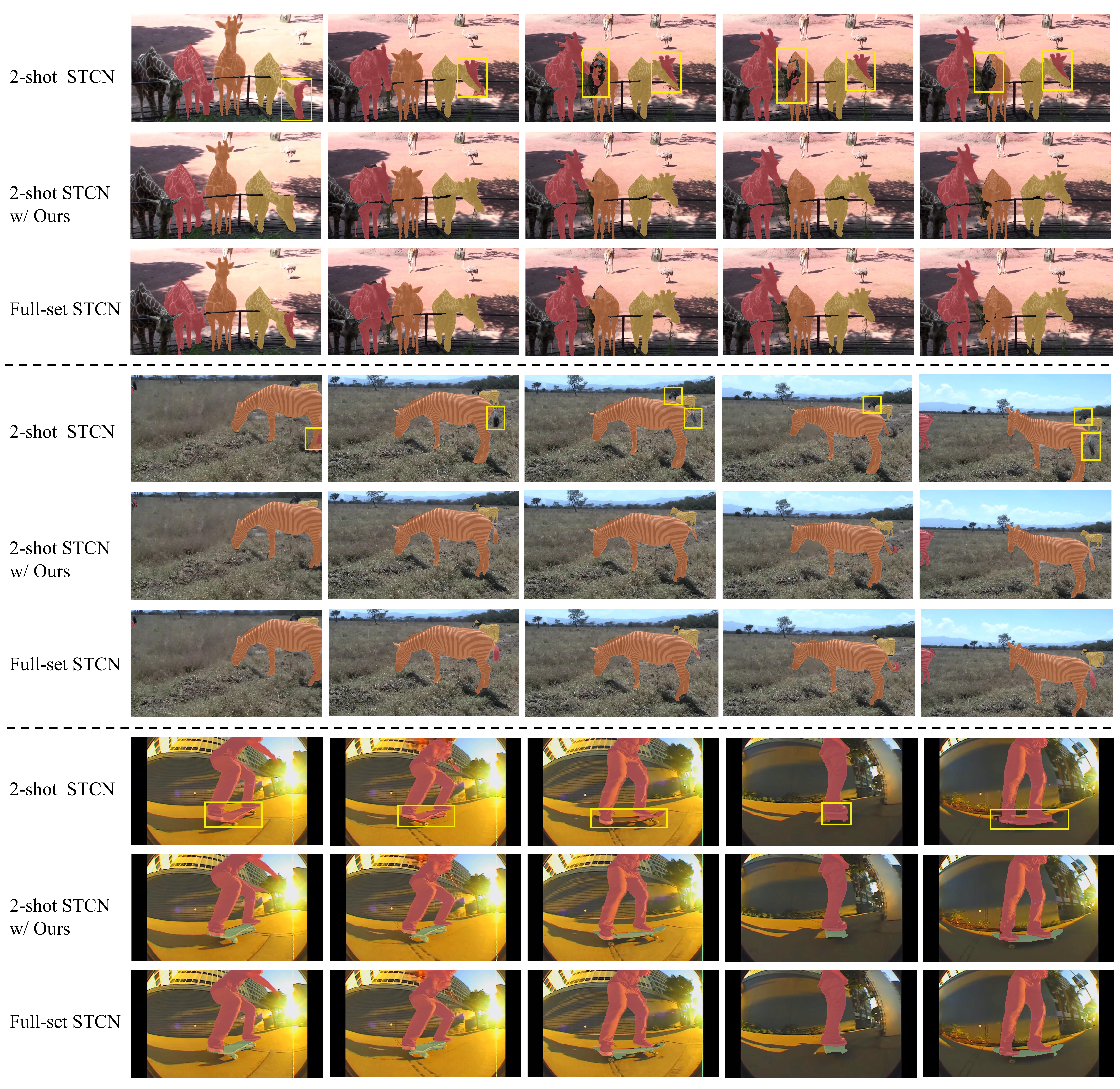}
  \caption{Mask prediction visualizations of naive 2-shot STCN, 2-shot STCN equipped with our approach, and STCN trained on fully labeled set. The yellow boxes highlight the regions where the naive 2-shot STCN model performs worse. The whole videos can be found in the \textit{``video''} folder in our supplemental package.}
  \label{fig:mask_vis}
\end{figure*}

\noindent \textbf{Visualization of mask prediction.} 
As shown in \cref{fig:mask_vis}, we visualize the mask predictions on three testing videos for several STCN variants: naive 2-shot STCN, 2-shot STCN with our training paradigm, and full-set STCN. The naive 2-shot STCN predicts pleasant masks, but fails to handle local details (\textit{e.g.} zebra tail in the second video). Please refer to the full video results in the \textit{``video''} folder in our supplemental material package.

\section{Limitations and future work}
\label{sec:limi}
Our 2-shot VOS methodology still shows a slight performance drop when applying to the methods that require more input frames during training~(\textit{e.g.} XMem~\cite{cheng2022xmem}). We will focus on addressing the problem of error propagation during pseudo-labeling in the future.

\begin{figure}[t]
  \centering
  \includegraphics[width=0.95\columnwidth]{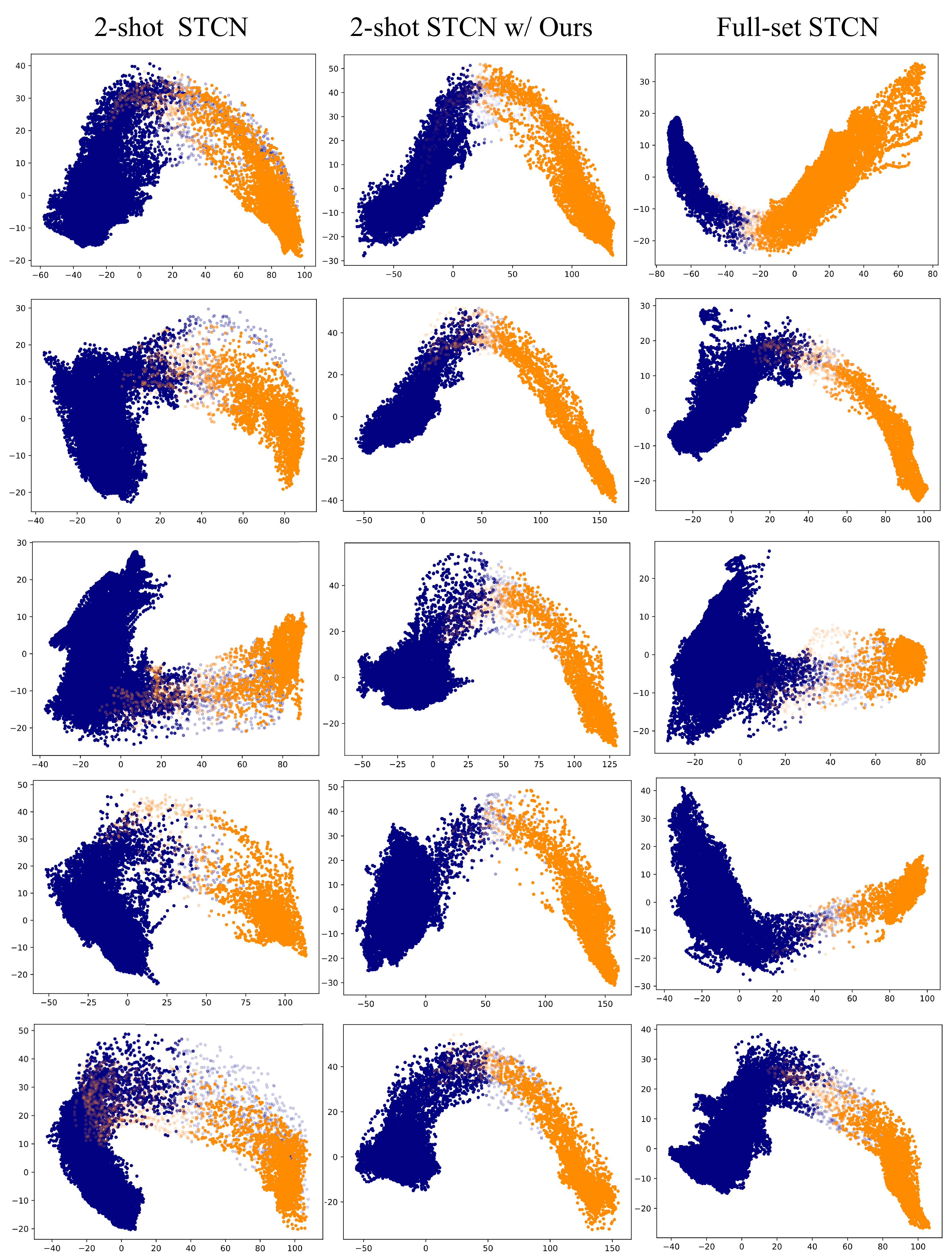}
  \caption{Feature space visualization with PCA. We compare naive 2-shot STCN, 2-shot STCN equipped with our approach, and full-set STCN~\cite{cheng2021rethinking}. Orange:
  foreground; blue: background.}
  \label{fig:feature_vis}
\end{figure}

\end{document}